%% file: main.tex
\documentclass[11pt]{article}

\usepackage[margin=1in]{geometry}

\usepackage[T1]{fontenc}
\usepackage{graphicx}
\usepackage{fancyhdr}
\usepackage{caption}
\usepackage[authoryear,round]{natbib}
\fancypagestyle{firstpage}{%
  \fancyhf{}%
  \fancyhead[L]{\includegraphics[width=1.15in]{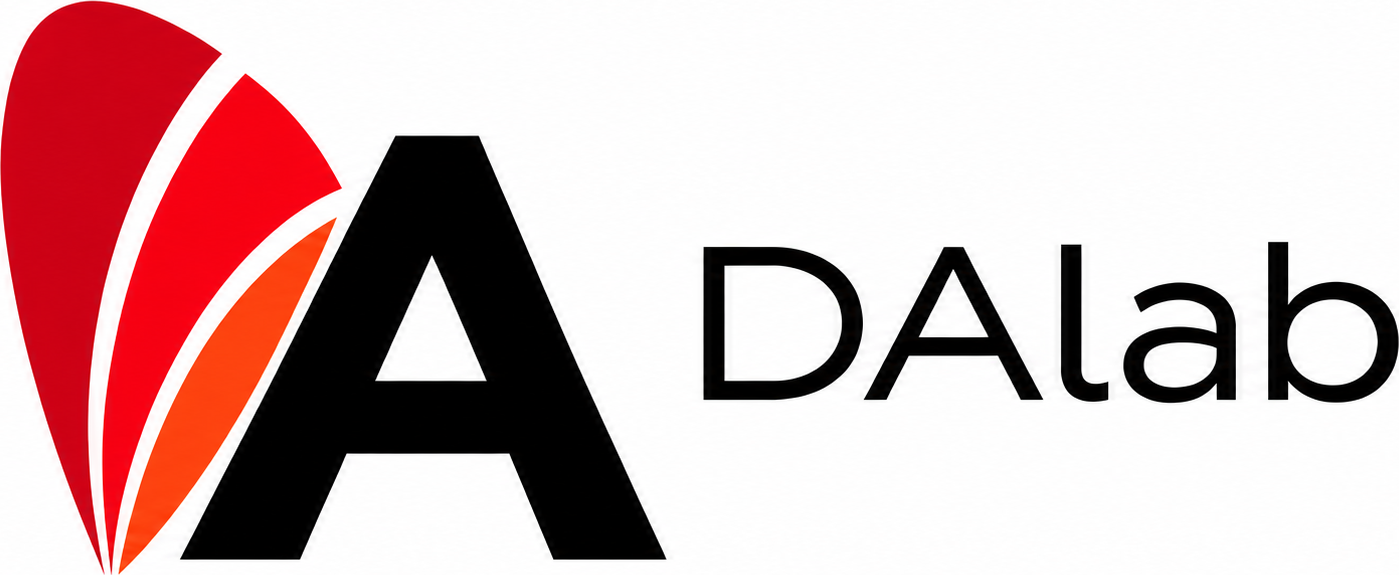}}%
  \fancyhead[R]{\small November 24, 2026}%
  \fancyfoot[L]{\raisebox{48pt}[0pt][0pt]{\begin{minipage}[b]{0.9\textwidth}
    \hrule height 0.4pt\vspace{3pt}
    \scriptsize
    \textbf{Contact information:}\\
    \texttt{\{jiahaorui, quzihan1, meiziyang1, zhaodousheng,}\\
    \texttt{zhangjiahao125, lihan158\}@huawei.com}\\
    \texttt{dongyundi206@gmail.com} \textperiodcentered\ \texttt{xgqzhangchi@126.com}\\
    \texttt{zzzyesyes@outlook.com} \textperiodcentered\ \texttt{magolorcz@gmail.com}
  \end{minipage}}}%
  \fancyfoot[R]{\thepage}%
}
\usepackage{placeins}
\usepackage{amsmath,amssymb}
\usepackage{booktabs,tabularx}
\usepackage{microtype}
\usepackage{titlesec}
\usepackage{xcolor}
\usepackage{xurl}
\definecolor{paperred}{HTML}{7D1F2A}
\usepackage[colorlinks=true,linkcolor=paperred,citecolor=paperred,urlcolor=paperred]{hyperref}
\makeatletter
\newenvironment{resultfigure}{%
  \begin{figure}[!htbp]\centering
}{\end{figure}}
\newenvironment{resulttable}{%
  \par\medskip\noindent\begin{minipage}{\columnwidth}%
  \def\@captype{table}\centering
}{\end{minipage}\par\medskip}
\makeatother

\titleformat{\section}
  {\normalfont\Large\bfseries\color{paperred}\raggedright}
  {\thesection}{1em}{}

\titleformat{\subsection}
  {\normalfont\large\bfseries\color{paperred}\raggedright}
  {\thesubsection}{1em}{}

\title{
\color{paperred}\bfseries RubikSQL-HB: Agentic Text-to-SQL\\
with Knowledge in the Loop
}

\author{
Haorui Jia$^{1,*}$ \quad Zihan Qu$^{1,*}$ \quad Yundi Dong$^{1,*}$ \\
Ziyang Mei$^{1}$ \quad Chi Zhang$^{1}$ \quad Zezhi Zheng$^{3}$ \quad Dousheng Zhao$^{1}$ \\
Jiahao Zhang$^{1,\dagger}$ \quad Zui Chen$^{2,\dagger}$ \quad Han Li$^{1,\dagger}$ \\[6pt]
\small $^{1}$Huawei Technologies Co., Ltd. \quad $^{2}$DeepSeek-AI \quad $^{3}$Alibaba Group \\[4pt]
\small $^{*}$Equal contribution \quad $^{\dagger}$Corresponding authors
}

\date{}

\begin{document}

\raggedbottom
\maketitle
\thispagestyle{firstpage}

\input{body/overview-figure}
\clearpage
\input{body/abstract}
\input{body/introduction}
\input{body/related-work}
\input{body/overview}
\input{body/retrieval-figure}
\input{body/method}
\input{body/author-summary-table}
\input{body/protocol}
\input{body/results}
\input{body/intervention-analysis}
\input{body/evolution-analysis}
\input{body/discussion}
\input{body/limitations}
\input{body/conclusion}

\begingroup
\small
\bibliographystyle{plainnat}
\bibliography{references}
\endgroup

\clearpage
\appendix
\input{appendix/reproducibility}

\end{document}

%% file: body/overview-figure.tex
\begin{center}
\begin{minipage}{\textwidth}
    \centering
    \includegraphics[width=\linewidth]{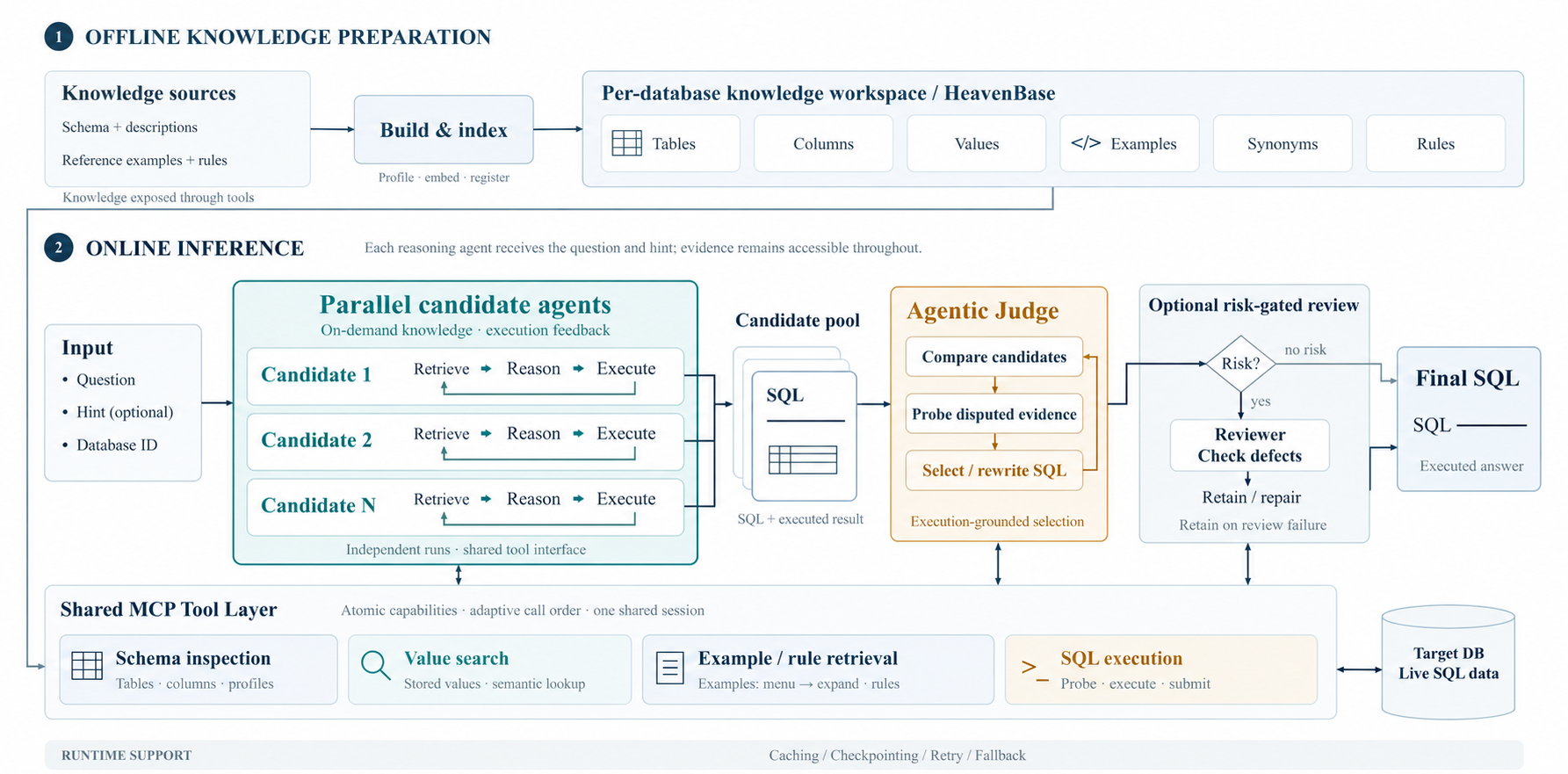}
    \captionof{figure}{RubikSQL-HB overview. Offline preparation organizes a
    per-database knowledge workspace; online Candidates query that resource
    within their retrieval--reasoning--execution loops. The Judge reuses the same
    interfaces, followed by optional risk-gated review.
    Bidirectional links denote tool access under each agent's permissions;
    activities inside agent boxes do not prescribe a fixed call sequence.}
    \label{fig:overview}
\end{minipage}
\end{center}

%% file: body/abstract.tex
\begin{abstract}
Database agents need more than schema structure: business meanings,
stored values and reference queries supply knowledge that becomes relevant
at different points in reasoning. We introduce RubikSQL-HB, which exposes
these materials through queryable, database-specific knowledge
workspaces built on the HeavenBase platform.
Its knowledge interface remains available as an agent constructs and
investigates an answer.
Candidate agents interleave knowledge access with SQL construction and
execution; a downstream Judge can reuse the same interfaces to investigate
disagreements. A two-stage example interface separates discovering a
reference question from expanding its SQL.
On a 500-question BIRD Mini-Dev evaluation, the full pipeline reaches
75.2\% execution accuracy, against 71.2\% for
first-candidate selection
and 70.4\% for execution-majority voting, with a correct candidate
available on 79.0\% of questions. The raw-knowledge condition, which
also regenerates candidates without the built workspace, is 3.0
points lower, and adjudication with tools and revision accounts for
most of the gain over selection baselines.
A recorded execution
probe shows how continued investigation resolves a candidate disagreement.
Separate archived runs show semantic revision recovering answers absent
from their original candidate pools. A versioned development study further
illustrates how recurring corrections become reusable execution checks.
The system and results connect knowledge-interface design to agent
decision quality: queryable knowledge supports the construction of useful
alternatives, and evidence-aware adjudication connects those alternatives
to a final answer. Keeping knowledge in the loop lets agents acquire
relevant observations as their reasoning develops.
\end{abstract}

%% file: body/introduction.tex
\section{Introduction}
\label{sec:introduction}

A database schema describes structure, but answering a business question
also requires interpreting that structure. An agent may need to identify
the stored code for a business term, distinguish an event table from an
entity table, or inspect a reference query's aggregation pattern.
Documentation and observed values supply different parts of this
knowledge. Making these materials usable during inference is a central
design problem for database agents.

Knowledge reaches an agent's reasoning in two complementary forms.
The \emph{workspace} holds stable, reusable background knowledge:
documented meanings, value profiles, synonyms, rules and reference
queries, organized per database and consulted on demand. The
\emph{execution probe} supplies the facts of the current hypothesis:
running a query over the live database and observing what comes back.
The first answers ``what does this field mean''; the second answers
``what does the data actually contain under this reading''. RubikSQL-HB
keeps both accessible throughout reasoning, so an agent can move
between interpretation and observation as its query develops.

The relevant knowledge is not always apparent at the start of a task.
A first query can reveal an unfamiliar value or an unexpected multiplicity;
two plausible answers can expose a disagreement about the quantity being
counted. An initial context gives the agent a starting point, while later
observations create more specific information needs. Database reasoning
therefore benefits from an interface through which the agent can return
to the underlying knowledge as its query develops.

RubikSQL-HB organizes this access around a persistent, database-specific
knowledge workspace. The central design choice is to make knowledge access
an action within reasoning, so agents can query relevant evidence as their
information needs develop. Built on HeavenBase, the workspace connects schema
descriptions to selected content and reference knowledge. RubikSQL-HB
defines the domain entities and their construction, then exposes atomic
tools for inspecting them. A Candidate can retrieve a column profile,
search a stored value or examine a query example within the same loop
that constructs and executes SQL. HeavenBase supplies the generic data
and workspace infrastructure; RubikSQL-HB supplies its database reasoning
interfaces and policies.

The Candidate is the core online process. It chooses its next observation
from the question and previous tool responses, updates its context, and
submits SQL for execution. Several Candidates can use the same knowledge
services with separate conversational histories. A downstream Judge
receives their SQL and execution summaries, then uses the same interfaces
to investigate a disagreement or revise an answer. An optional Reviewer
extends this process to selected outputs. Knowledge access thus remains
available across stages, without requiring agents to exchange their entire
reasoning histories.

Within this architecture, interface design determines the decisions an
agent can make about knowledge. RubikSQL-HB separates example discovery
from SQL expansion: a short menu exposes reference questions, and the
agent chooses which SQL to inspect. This adds a decision about information
access before the reference query enters context. Prior work studies
example organization~\citep{Gao2024text}, persistent knowledge for SQL
reasoning~\citep{Chen2025rubiksql}, and tool-assisted
refinement~\citep{Talaei2024chess}. Our system connects these concerns through
a shared, queryable workspace used during both query construction and
adjudication.

We evaluate the design at these interfaces, examining knowledge access,
candidate construction, and downstream decision policies on a
500-question BIRD Mini-Dev subset. The full pipeline reaches 75.2\%
execution accuracy, versus 71.2\% for first-candidate selection and
70.4\% for execution-majority voting; a correct candidate is available
on 79.0\% of questions, so adjudication and revision recover about
half of the gap that selection alone leaves open. Fixed-pool
permission comparisons and a targeted retest examine the
policies that direct further investigation and semantic correction.

Execution experience also supplies material for system improvement.
Recurring errors in aggregation grain, denominator choice or added joins
can be translated into narrowly scoped execution checks. A supplementary
versioned development study follows this process from diagnosis to
adoption. It extends the analysis of within-task reasoning to the
reuse of specific corrections in later versions.

The report makes three connected contributions:
\begin{itemize}
  \item A database knowledge interface for iterative reasoning. Typed
  records and two-stage example access let Candidates choose relevant
  observations during SQL construction. HeavenBase supplies the generic
  workspace infrastructure; RubikSQL-HB defines the domain knowledge and
  its model-facing interfaces (Section~\ref{sec:knowledge-workspace}).
  \item An evidence-aware decision layer that reuses these interfaces
  after generation. The Judge can compare existing evidence, investigate
  a disagreement and revise SQL. Fixed-pool comparisons separate the
  contribution of adjudication from additional observation and revision
  permissions, which together account for a 4.2-point gain over
  execution-majority voting (Sections~\ref{sec:stage-reuse}
  and~\ref{sec:component-ablations}).
  \item An empirical analysis of knowledge use and semantic decisions.
  Stage-level outcomes connect candidate coverage to final answers.
  A supplementary development study follows recurring corrections
  into targeted execution checks
  (Sections~\ref{sec:candidate-coverage} and~\ref{sec:evolution}).
\end{itemize}
Together, these contributions trace one design principle through a
complete knowledge lifecycle: a workspace accumulates database
knowledge before inference; agents consult it, and the live database,
while constructing and adjudicating SQL; and recurring corrections are
consolidated back into reusable checks. Keeping knowledge available at
each point of decision is what we refer to as knowledge in the loop.

%% file: body/related-work.tex
\section{Background and Related Work}
\label{sec:related}

Spider evaluates complex queries across unseen
schemas~\citep{Yu2018spider}, while BIRD emphasizes database contents,
external knowledge, and execution-based assessment~\citep{Li2023already}.
Spider~2.0 extends evaluation to enterprise workflows with large schemas,
multiple dialects, and project context~\citep{Lei2024spider}. RubikSQL-HB
addresses database-grounded reasoning in the BIRD setting.

\paragraph{Grounding and query construction.}
Schema interpretation and query structure are distinct sources of error.
RAT-SQL models schema relations~\citep{Wang2020relation}, and BRIDGE
incorporates relevant cell values into joint question--schema
representations~\citep{Lin2020bridging}. IRNet generates through an
intermediate representation~\citep{Guo2019towards}; RESDSQL separates schema
ranking from skeleton generation~\citep{Li2023resdsql}. SmBoP composes
program subtrees bottom-up~\citep{Rubin2021smbop}, whereas PICARD constrains
decoding through incremental parsing~\citep{Scholak2021picard}. These
mechanisms improve query formation but do not equate syntactic validity
with intended meaning. RSL-SQL addresses the risk of dropping necessary
schema elements through bidirectional linking and full/simplified schema
selection~\citep{Cao2024robust}. This risk motivates keeping omitted
evidence retrievable during subsequent reasoning.

\paragraph{Context and persistent knowledge.}
DIN-SQL combines task decomposition with
self-correction~\citep{Pourreza2023decomposed}. DAIL-SQL studies prompt
representation and example selection and organization~\citep{Gao2024text};
C3 combines clear prompting, calibration hints, and output
consistency~\citep{Dong2023zero}. Beyond prompt construction, RubikSQL
maintains an agentic knowledge lifecycle for generation and
refinement~\citep{Chen2025rubiksql}, and MaP-SQL retrieves memory for
training-free listwise selection~\citep{Jeong2026replacing}. RubikSQL-HB's
HeavenBase workspace exposes database evidence within the Candidate's
query-construction loop and keeps it accessible to the Judge. Its
reasoning interfaces connect database-scoped knowledge, staged example
discovery and expansion, and continued evidence access after generation.

\paragraph{Agents and candidate selection.}
ReAct interleaves reasoning and actions to acquire
observations~\citep{Yao2022react}. MAC-SQL assigns schema selection,
decomposition, and execution-feedback refinement to collaborating
agents~\citep{Wang2025mac}; MAG-SQL uses soft schema linking and iterative
sub-SQL refinement~\citep{Xie2024multi}. ReFoRCE combines self-refinement
and consensus with execution-guided column
exploration~\citep{Deng2025reforce}. Active evidence gathering is therefore
an established approach. CHASE-SQL pairs multi-path generation with
preference-optimized selection~\citep{Pourreza2024chase}, while XiYan-SQL
combines multiple generators and selection~\citep{Liu2026xiyan}. RubikSQL-HB
also separates these stages, with a tool-assisted Judge that can retrieve
further evidence or produce SQL outside the pool. Its evaluation separates
candidate coverage from successful selection.

\paragraph{Execution feedback and verification.}
Execution-guided decoding rejects faulty partial
programs~\citep{Wang2018robust}. CHESS combines retrieval, revision, and
unit-test-based checking~\citep{Talaei2024chess}; DPC compares SQL and Python
interpretations on synthesized distinguishing
databases~\citep{Li2026training}; PV-SQL combines database probing with
rule-based verification~\citep{Tian2026synergizing}. RubikSQL-HB uses
target-database observations and workspace knowledge, followed by a
heuristically gated Reviewer. These observations support decisions but
provide no semantic guarantee: different queries can agree on one database
instance. Distilled test suites approximate semantic equivalence across
multiple instances~\citep{Zhong2020semantic}, illustrating the limits of
single-instance execution checks.

%% file: body/overview.tex
\section{RubikSQL-HB Overview}
\label{sec:overview}

RubikSQL-HB receives a target database, a natural-language question and an
optional domain hint, and returns SQL with its execution result.
Figure~\ref{fig:overview} separates offline knowledge preparation from the
online reasoning process. Queryable knowledge connects construction and
adjudication: an agent can return to the same resource when execution
or candidate disagreement exposes a new information need.
The workspace persists across requests;
individual agent conversations belong to their own inference runs.

Offline preparation organizes six domain knowledge types: tables, columns,
selected values, examples, synonyms and rules. Database-level overview
metadata complements these records. Documentation and profiles explain
the target schema, while transferred examples provide reference usage.
HeavenBase supplies the underlying Entity/Workspace and storage/indexing
abstractions. RubikSQL-HB defines the contents and database scope of these
entities, along with the tools through which agents read them. SQL
execution remains connected to the external target database.

Online, a Candidate alternates knowledge access with query construction
and execution. The shared Model Context Protocol (MCP) interface supports
schema inspection and content lookup as separate actions. Example
discovery can precede selective SQL expansion. Tool observations enter
the Candidate's context, allowing its next call to respond to information
learned during the task.

Parallel Candidates conduct separate searches over this shared resource.
Their SQL and execution summaries form the input to the Judge, which can
continue gathering evidence before selecting or revising an answer.
A risk-gated Reviewer optionally inspects the result. This organization
reuses knowledge and execution capabilities across stages while keeping
each agent's conversational state separate. Checkpoints and verified-query
fallbacks preserve completed work when inference is interrupted.

Execution traces also support development between versions. A separate
evaluation loop records proposed policy changes and compares each child
with its parent before updating the baseline. Section~\ref{sec:evolution}
examines this process. It governs the configuration used by subsequent
runs, while the online workflow in Figure~\ref{fig:overview} queries the
prepared knowledge workspace.

%% file: body/retrieval-figure.tex
\begin{figure}[!t]
    \centering
    \includegraphics[width=0.86\textwidth]{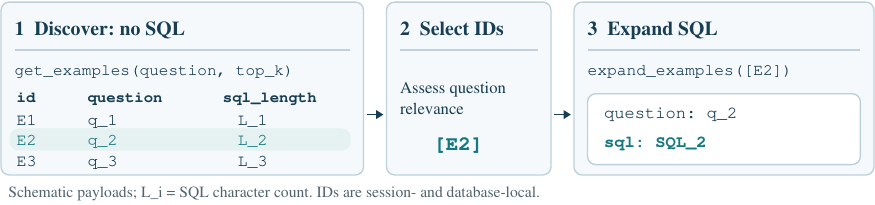}
    \caption{Two-stage example retrieval. Discovery exposes only a
    session-local ID, question, and SQL character count; selected IDs resolve
    to retained question--SQL pairs in the same session and database.
    The agent chooses which reference SQL enters context. IDs and payloads
    are schematic.
    Candidate coverage and final accuracy under this interface are
    evaluated separately in Section~\ref{sec:component-ablations}.}
    \label{fig:two-stage-retrieval}
\end{figure}

%% file: body/method.tex
\section{Queryable Knowledge for Autonomous SQL Reasoning}
\label{sec:method}

RubikSQL-HB prepares database knowledge for use within an agent's reasoning
loop. A Candidate queries this structured resource while constructing SQL;
later agents reuse its interfaces to inspect a completed answer.

\subsection{Building the Database Knowledge Workspace}
\label{sec:knowledge-workspace}

The workspace combines information that answers different kinds of
database questions. Schema metadata locates a field, while documentation
explains its meaning. Profiles add observations about stored content,
such as value formats and null frequencies. Reference queries illustrate
usage, and rules express reusable interpretation guidance.
Table~\ref{tab:knowledge-sources} relates these sources to their online use.
Database-level metadata also provides an overview of the target schema.

\begin{resulttable}
  \small
  \setlength{\tabcolsep}{3pt}
  \begin{tabularx}{\columnwidth}{>{\raggedright\arraybackslash}X>{\raggedright\arraybackslash}X}
    \toprule
    Knowledge and source & Reasoning use \\
    \midrule
    Tables / columns: schema, documentation and profiles &
      Locate fields, inspect keys and interpret granularity \\
    Values: selected categorical content &
      Match business expressions to stored literals \\
    Examples: transferred training question--SQL pairs &
      Inspect reference query patterns \\
    Synonyms: vocabulary-to-schema mappings &
      Connect user wording to canonical identifiers \\
    Rules: prepared interpretation guidance &
      Check a proposed reading of the question \\
    \bottomrule
  \end{tabularx}
  \caption{Sources and uses of the six domain knowledge types.
  Database-level overview metadata is stored separately.}
  \label{tab:knowledge-sources}
\end{resulttable}

Preparation first registers schema metadata and attaches documentation.
Profiling derives statistics from the target database and stores generated
descriptions alongside the documented meaning. Example construction adapts
training-source question--SQL pairs to the target schema. The evaluated
snapshot selects source examples using the fixed evaluation questions and
hints, without supplying target gold SQL to the transfer model; the
construction record is in Appendix~\ref{app:run-record}. The resulting
examples and interpretation guidance are stored before inference.

HeavenBase provides the generic \texttt{Entity} and \texttt{Workspace}
abstractions and the storage/indexing capabilities. RubikSQL-HB defines
the database-specific entity fields, prepares their contents, and exposes
the retrieval and SQL tools used by its agents. Scalar records and vector
indexes support complementary access paths. Knowledge records carry a
database identifier, and database-specific retrieval restricts that scope
before similarity ranking. The target database remains the execution
backend; the workspace stores its descriptions and selected content.

The tool interface preserves distinctions between sources. An agent can
inspect a column profile without retrieving an example, or search stored
values without accepting an example's join pattern. Rules are retrieved
through title and descriptive content, with literal triggers supplying
applicable guidance during a conversation. Synonym notes augment the
question while preserving its original wording. These mechanisms make
knowledge accessible in small, task-relevant units. A returned rule is
guidance to assess against the question and database, not an execution
constraint enforced by the backend.

Example access further separates discovery from expansion
(Figure~\ref{fig:two-stage-retrieval}). \texttt{get\_examples} returns a
menu containing a session-local identifier, reference question and SQL
length. \texttt{expand\_examples} resolves selected identifiers to complete
question--SQL pairs in that database and session. The agent can inspect
the reference question before deciding to add its SQL to context.
Section~\ref{sec:component-ablations} tests the consequences of this
interface for candidate coverage and final accuracy.

\begin{figure}[!t]
    \centering
    \includegraphics[width=\textwidth]{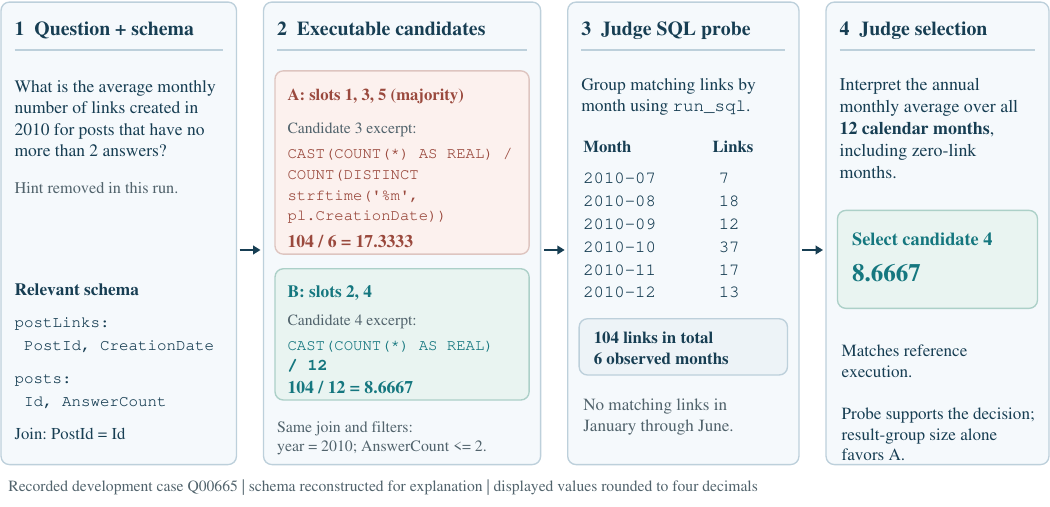}
    \caption{A recorded denominator dispute (development question Q00665,
    \texttt{codebase\_community}). All five candidates execute successfully;
    three return $104/6$ and two return $104/12$. The Judge's monthly-count
    probe exposes the six observed months; the question asks for a full-year
    average. Candidate 4 matches the reference. SQL excerpts share the join
    and filters; displayed values are rounded and schema is reconstructed.
    This qualitative execution-probe case is documented in
    Appendix~\ref{app:execution-case}.}
    \label{fig:monthly-links-case}
\end{figure}

\subsection{Candidate Reasoning with On-Demand Evidence}
\label{sec:candidate-loop}

A Candidate begins with the question and optional hint, its tool
definitions, and the configured reasoning instructions. The runtime adds
matching synonym notes and triggered rules. The agent then selects
observations as its interpretation develops. It can request database or
table information to choose a join path, inspect a field's profile to
resolve its meaning, or retrieve an example for a query pattern. These
calls are interleaved with SQL construction and execution rather than
performed as one fixed retrieval stage.

Each tool response enters that Candidate's own conversation. Schema and
knowledge tools return records; SQL tools return column information and
bounded row previews, or an error that the agent can use to revise its
query. The runtime retains the full execution result separately from the
model-visible rendering. A new observation can change the next action:
an unexpected result may prompt a value lookup, and a discovered
one-to-many relation may prompt a distinct-count probe. The agent can
therefore return to the workspace after it has begun executing SQL.

\paragraph{A recorded knowledge-to-query sequence.}
In retrieval-study r2/Q00532, the question asks for the display names of
users whose accounts were created in 2011. Candidate slot zero first
requests database information, then calls \texttt{get\_col\_info} on
\texttt{users.CreationDate}. The returned profile identifies account
creation time, reports text storage, and gives timestamp-format examples.
The Candidate subsequently executes:
\begin{quote}
\footnotesize\ttfamily\raggedright
SELECT DisplayName FROM users\\
WHERE strftime('\%Y', CreationDate) = '2011'
\end{quote}
The saved execution returns 4,430 rows and the adopted candidate score is
correct. This sequence shows a specific field-level knowledge response
entering a successful query-construction process within the Candidate
loop. Appendix~\ref{app:knowledge-case}
records the trace and scoring provenance.

Submission commits self-contained SQL over the base database and obtains
its execution result. An already-verified identical statement can reuse
that result. The implementation requests a final output check before
accepting the submission when runtime limits permit. Turn and time guards
bound exploration; a successful base-query verification can supply a
fallback if explicit final submission does not complete. A run with no
submitted or salvageable query is an inference failure. These completion
rules distinguish generating plausible SQL text from returning an answer
that the database has actually produced.

\subsection{Reusing Evidence Access across Stages}
\label{sec:stage-reuse}

For a question $q$, optional hint $h$, database $D$ and knowledge
workspace $K_D$, independent Candidate agents produce SQL $s_i$ with
execution result $r_i$. Their pool is
\begin{equation}
    \mathcal{C}(q,D)=\{(s_i,r_i)\}_{i=1}^{N}.
\end{equation}
The evaluated system uses five candidates. They share the underlying
knowledge and execution services, but each keeps its own conversation and
run-scoped references. Sharing infrastructure gives every Candidate access
to the same database resources without merging their interpretations.

The Judge receives candidate SQL and execution summaries, including row
counts, bounded previews and errors. Result-group labels and sizes expose
agreement on the current database instance. The handoff contains these
artifacts rather than the Candidates' conversational histories. When a
semantic assumption becomes disputed, the Judge can revisit the workspace
or execute a targeted probe through the same tool interface.

Figure~\ref{fig:monthly-links-case} illustrates this continued access.
Three Candidates average monthly links over the six observed months,
while two divide by the full twelve-month year. The Judge's monthly-count
probe reveals the observed period; the question supplies the intended
full-year denominator. The Judge selects a correct minority candidate.
This case is settled by execution evidence rather than workspace
records: the dispute concerns the data's actual contents, so the
decisive observation comes from a live probe rather than a stored
profile. Disputes about meaning, such as what a field denotes, draw
instead on workspace knowledge, as the column-profile trace in
Section~\ref{sec:candidate-loop} shows. Both evidence types reach the
Judge through the same interface.

The Judge can select an original candidate or submit revised SQL outside
the pool, according to its permissions. Selection uses the alternatives
already available; revision allows the Judge to construct an answer
beyond them. The fixed-pool evaluation separates observation-tool access
from revision permission to measure these decisions without changing
generation.

A Reviewer is an optional final use of the same reasoning machinery.
The evaluated risk gate uses coarse SQL/output signals, including empty
results, null cells and selected filter or expression patterns. A triggered
Reviewer can retain or revise the Judge's answer; review failure preserves
that answer. Comparing no review, unconditional review and gated review
separates the activation policy from the resulting SQL edits.
This extension uses the shared evidence interfaces without making
review a prerequisite for Candidate reasoning.

\FloatBarrier

%% file: body/author-summary-table.tex
\begin{table}[t]
\centering
\small
\setlength{\tabcolsep}{5pt}
\renewcommand{\arraystretch}{1.1}
\caption{Main ablation on the 500-question evaluation cohort.
Full-M (RG) is the primary configuration; other rows vary one
permission, model, or knowledge-access factor at a time relative to it.
Candidate-pool references and permission rows share one candidate
generation; the knowledge-access rows (G0/G1/K0) regenerate their own
pools per strategy.}
\label{tab:author-ablation-summary}
\begin{tabular*}{\textwidth}{@{\extracolsep{\fill}}l l r r@{}}
\toprule
\textbf{Configuration} & \textbf{Description} & \textbf{Correct/500} & \textbf{EX (\%)} \\
\midrule
\multicolumn{4}{@{}l}{\textit{Candidate-pool references (no Judge)}} \\
Oracle & At least one of five candidates correct & 395 & 79.0 \\
FIRST & Always select the first candidate & 356 & 71.2 \\
MV & Execution-result majority vote & 352 & 70.4 \\
\addlinespace[3pt]
\multicolumn{4}{@{}l}{\textit{Judge permissions and review (qwen3.8-max Judge)}} \\
J00-M & Selection only; no tools, no revision & 358 & 71.6 \\
J01-M & Revision permission, no observation tools & 369 & 73.8 \\
J10-M & Observation tools, selection only & 370 & 74.0 \\
J11-M-R0 & Tools + revision, no review & 373 & 74.6 \\
J11-M-RA & Tools + revision, review all outputs & 372 & 74.4 \\
\textbf{Full-M (RG)} & \textbf{Tools + revision, risk-gated review (primary)} & \textbf{376} & \textbf{75.2} \\
\addlinespace[3pt]
\multicolumn{4}{@{}l}{\textit{Judge-model comparison (qwen3.7-plus Judge)}} \\
J00-P & Selection only; no tools, no revision & 351 & 70.2 \\
J11-P-R0 & Tools + revision, no review & 365 & 73.0 \\
Full-P & Tools + revision, risk-gated review & 368 & 73.6 \\
\addlinespace[3pt]
\multicolumn{4}{@{}l}{\textit{Knowledge-access settings (primary configuration)}} \\
G1-Full & Single-stage example retrieval & 374 & 74.8 \\
G0-Full & Example-entity access disabled & 371 & 74.2 \\
K0-Full & Raw knowledge base (no built workspace) & 361 & 72.2 \\
\bottomrule
\end{tabular*}
\end{table}

%% file: body/protocol.tex
\section{Experiments}
\label{sec:experiments}
\label{sec:results}

We evaluate the complete pipeline, then compare knowledge-access designs
and downstream decision policies. The evaluation is organized in four
parts with distinct roles. The main ablation
(Table~\ref{tab:author-ablation-summary}) gives the system-level
comparison: candidate-pool references, Judge permissions and review
policies, a Judge-model comparison, and knowledge-access settings.
Paired studies then attribute effects at fixed candidate pools: the
decision study tests Judge permissions and review on identical
candidates, and the retrieval study regenerates pools for each
example-access strategy. Archived analyses (candidate mixtures,
mechanism interventions, and rescored release runs) document how
historical configurations behave on their own saved pools, and the
versioned development study records how policies changed between
releases. Absolute scores are compared only within each part; the
record-level analyses of Sections~\ref{sec:candidate-coverage}
and~\ref{sec:evolution} use their own fixed pools and are kept separate
from the main ablation scores.

\subsection{Experimental Setup}
\label{sec:protocol}

We evaluate on the cleaned 500-question BIRD Mini-Dev development set
(\texttt{minidev\_500\_20251106}), spanning 11 databases. Five
\texttt{qwen3.7-plus} candidates answer each question; the primary
Judge and Reviewer use \texttt{qwen3.8-max} (Max), and the
model-comparison family substitutes \texttt{qwen3.7-plus} as the Judge
(Plus). Inference inputs contain the question and its hint or
evidence, but no gold SQL; gold queries are used only for scoring. All
configurations share the same questions and knowledge snapshot, with
\texttt{text-embedding-v3} retrieval. The decision study holds
candidate pools fixed to test Judge permissions and review; the
retrieval study generates separate pools for each example-access
strategy. Comparisons are paired within each study; their Full runs are
not pooled.

Execution accuracy (EX) measures agreement with the reference query's
execution result, scored over 500 questions per configuration. Terminal
inference failures count as incorrect. Candidate Oracle EX measures the
fraction of questions with at least one correct candidate:
\begin{align}
\mathrm{EX} &= \frac{100}{M}\sum_{j=1}^{M}E_j(s_j^{\mathrm{final}}),
\label{eq:execution-accuracy}\\
\mathrm{Oracle\ EX} &= \frac{100}{M}\sum_{j=1}^{M}
\mathbf{1}\!\left[\exists s\in\mathcal{C}_j^{\mathrm{SQL}}:E_j(s)=1\right].
\label{eq:candidate-oracle}
\end{align}
Here $M=500$, $E_j$ is the execution-match indicator, and
$\mathcal{C}_j^{\mathrm{SQL}}$ contains the five candidate slots,
including failed slots. A Judge that writes new SQL can exceed
candidate coverage, so Oracle EX is not a bound on a revision-enabled
system.

Record-level analyses (coverage-to-answer transitions, candidate-count
sweeps, the discordant-unit retest) resample questions and combine
repetitions where applicable; transition totals are reported as
question--repetition records. Paired 95\% bootstrap intervals resample
questions 10,000 times. Repairs and regressions count
incorrect-to-correct and correct-to-incorrect changes.
Appendix~\ref{app:reproducibility} records the recovery history and
detailed configuration; archived candidate-mixture and routing studies
retain their own scoring protocols. We also analyze a retest of 46
historically discordant units and independently rescore archived Judge
runs. These diagnose decision behavior on their own fixed pools;
neither changes any cohort's predictions or denominator.

Section~\ref{sec:evolution} separately tracks policy development on an
older fixed 330-question cohort with three Candidates. Its version
comparisons retain their own cohort and configuration.

%% file: body/results.tex
\subsection{Main Results}
\label{sec:main-results}

Table~\ref{tab:author-ablation-summary} reports the main ablation
results on the primary evaluation cohort. The primary configuration,
Full-M (RG) with the Max Judge and risk-gated review, reaches 75.2\% EX.
Candidate-pool references bound the decision layer: Oracle shows that a
correct candidate exists on 79.0\% of questions, while first-candidate
and execution-majority selection reach only 71.2\% and 70.4\%. The Judge
with tools and revision (J11-M-R0, 74.6\%) exceeds both selection
references, and the primary configuration reaches 75.2\%; these
gaps of 3--5 points (15--24 questions) are the clearest
contrasts in the table. Within the permission grid, both capabilities
raise EX over the passive Judge (J10-M 74.0\%, J01-M 73.8\%, versus
J00-M 71.6\%), and combining them yields the highest pre-review score
(J11-M, 74.6\%). The tools--revision interaction is
$(74.6-74.0)-(73.8-71.6)=-1.6$ points: each capability adds less when
the other is present, so the grid's clearest contrasts are the
3--4-point gains of any active permission over the passive Judge.
Review policies are compared on shared Judge outputs below. Under the
Plus Judge the same ordering holds at a lower level (70.2\% $\to$
73.0\% $\to$ 73.6\%), so the gated-review pipeline is 1.6 points more
accurate with the Max Judge. Knowledge access produces the largest
gap in this comparison: the K0 condition, which regenerates its
candidates under a raw knowledge base, is 3.0 points below the
primary configuration (72.2\% vs.\ 75.2\%); this contrast therefore
includes candidate-generation differences alongside knowledge access.
The two-stage example interface
(Full-M, 75.2\%) is 0.4--1.0 points above the single-stage variant
(G1-Full, 74.8\%) and disabled example access (G0-Full, 74.2\%).

The headline gap in Table~\ref{tab:author-ablation-summary} frames the
rest of the evaluation: a correct candidate exists on 79.0\% of
questions, but execution-majority selection returns 70.4\%. Better
knowledge access creates useful alternatives; converting them into
final answers is a separate decision. The following sections examine
that decision: how Candidates acquire knowledge during construction,
how the Judge extends the same access to adjudication, and how far the
coverage--answer gap closes
(Section~\ref{sec:candidate-coverage}).

\subsection{Component Ablations}
\label{sec:component-ablations}

\paragraph{Example access and retrieval interface.}
The retrieval study compares no external reference examples (G0), complete
examples returned in one stage (G1), and a menu followed by selective SQL
expansion (G2). Non-example knowledge remains available in every group.
G1 and G2 share the example index and ranking policy; each strategy applies
throughout generation, adjudication, and review.

\begin{resulttable}
  \centering
  \small
  \setlength{\tabcolsep}{3pt}
  \begin{tabular}{lr}
    \toprule
    Example strategy & Full EX (\%) \\
    \midrule
    G0: no examples & 74.2 \\
    G1: one stage & 74.8 \\
    G2: two stages & 75.2 \\
    \bottomrule
  \end{tabular}
  \caption{Example-access comparison on the primary evaluation cohort.
  All entries are EX percentages under the primary configuration.}
  \label{tab:retrieval-review-ablation}
\end{resulttable}

The two-stage interface (G2) attains the highest final EX in this
comparison, 1.0 points
above disabled example access (G0) and 0.4 points above one-stage
retrieval (G1); the interface gaps are small and are not interpreted
as a stable ordering. The stage-wise view connects the larger set of
available solutions to downstream selection: the Judge can draw on
more useful alternatives created through two-stage reference access.

The interface is available without being used uniformly. Across the
recorded repetitions, G1 exposes retrieved examples on 497 and 492
questions, while G2 exposes a menu on 435 and 411 questions and expands
SQL on 165 and 132. Agents choose whether to use the retrieved
references, so interface availability and actual expansion are
recorded separately.

\paragraph{Judge evidence and revision.}
Adjudication on existing evidence already pays off: the passive Judge
(J00-M, 71.6\%) exceeds execution-majority voting (70.4\%) in
Table~\ref{tab:author-ablation-summary}. Tool access extends this
decision layer to disputed
assumptions, as the denominator probe in Figure~\ref{fig:monthly-links-case}
illustrates; revision also permits a new answer when selection is
insufficient. We cross observation-tool access with permission to revise SQL
outside the pool. The four variants receive identical candidate SQL,
order, execution previews, and initial context; all run without a Reviewer.
Selection-only variants must choose an original candidate. Tools-off
variants cannot request additional observations.

\begin{resulttable}
  \centering
  \setlength{\tabcolsep}{3pt}
  \small
  \begin{tabular}{lc r}
    \toprule
    Judge & Tools / revision & EX (\%) \\
    \midrule
    J00-M & Off / Off & 71.6 \\
    J01-M & Off / On & 73.8 \\
    J10-M & On / Off & 74.0 \\
    J11-M & On / On & 74.6 \\
    \bottomrule
  \end{tabular}
  \caption{Fixed-pool Judge comparison before review, separating
  observation access from revision permission, on the primary
  evaluation cohort. J01 includes its recorded output fallback
  policy; see Appendix~\ref{app:run-record}.}
  \label{tab:judge-ablation}
\end{resulttable}

Additional tools change EX by $+2.4$ points under selection-only
permissions and $+0.8$ points when revision is allowed. Allowing
revision changes EX by $+2.2$ points without tools and $+0.6$ points
with tools. J11, which combines both permissions, has the highest
average EX in this comparison. The investigation and repair cases in
Sections~\ref{sec:stage-reuse} and~\ref{sec:candidate-coverage}
explain the additional capabilities' practical role: resolving a
specific semantic ambiguity and checking a proposed answer against it.
The average effects depend on the policy that decides when and how to
use those capabilities.

\begin{resultfigure}
  \includegraphics[width=0.56\columnwidth]{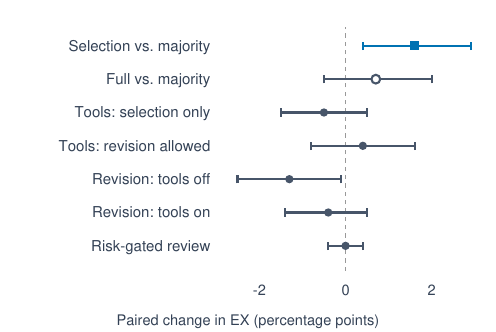}
  \caption{Paired permission effects in the two-repetition paired
  decision study (its own pools and scoring protocol; separate from
  the main ablation of
  Table~\ref{tab:author-ablation-summary}). Points denote mean EX
  differences; whiskers are question-clustered 95\% bootstrap
  intervals. The blue square isolates selection-only adjudication's
  gain over majority; other rows test Full and additional
  permissions.}
  \label{fig:decision-effects}
\end{resultfigure}

\paragraph{Semantic targets for adjudication.}
A targeted retest reruns all four arms on 46 question--repetition
units, covering 37 questions, selected because the arms disagree.
The candidate pools and initial Judge inputs remain fixed.
Table~\ref{tab:judge-retest} reports this selected subset separately
from the full-cohort scores.

\begin{resulttable}
  \centering
  \setlength{\tabcolsep}{7pt}
  \begin{tabular}{lrr}
    \toprule
    Judge & Original correct & Retest correct \\
    \midrule
    J00 & 30/46 & 25/46 \\
    J01 & 17/46 & 29/46 \\
    J10 & 25/46 & 27/46 \\
    J11 & 21/46 & 25/46 \\
    \bottomrule
  \end{tabular}
  \caption{Targeted retest on historically discordant units, not a new
  full-cohort score. Each arm reuses the original five-candidate pool.}
  \label{tab:judge-retest}
\end{resulttable}

J11's original five wins and fourteen losses against J00 become nine
wins and nine losses. Of the nineteen original differences, seven recur
in the same direction, eleven become ties and one reverses; ten new
differences appear elsewhere in the selected subset. These transitions
identify decisions that vary across fresh inferences.
The largest change, J01's twelve additional correct records, also depends
on output handling: all 46 new outputs carry a fallback tag, and 43 final
queries match an original candidate.

The recurring cases identify a more specific semantic issue. Adding
\texttt{DISTINCT} repairs patient-count queries Q01256 and Q01254, but
loses record-count queries Q01252 and Q01267; another recurring loss
omits a required non-null filter. The same SQL operation can help or hurt
depending on what the question counts. Aggregation grain and filter
semantics are therefore concrete targets for the Judge's investigation
and validation of a proposed correction. The retest
and its delivered evidence are documented in
Appendix~\ref{app:judge-retest}.

\paragraph{Reviewer activation.}
No review, review-all, and risk-gated review share the same J11 outputs.
The gated view reuses the same review result on triggered questions and
retains the Judge elsewhere.

\begin{resulttable}
  \centering
  \setlength{\tabcolsep}{5pt}
  \begin{tabular}{lr}
    \toprule
    Review policy & EX (\%) \\
    \midrule
    No review (J11-M-R0) & 74.6 \\
    Review all (J11-M-RA) & 74.4 \\
    Risk-gated (Full-M) & 75.2 \\
    \bottomrule
  \end{tabular}
  \caption{Reviewer effects on shared Judge outputs on the primary
  evaluation cohort. The gated view reuses the same review result on
  triggered questions and retains the Judge elsewhere.}
  \label{tab:reviewer-ablation}
\end{resulttable}

On shared Judge outputs, unconditional review scores 74.4\% and no
review 74.6\% (a one-question difference), while the risk-gated
configuration attains 75.2\% in this run. The gate concentrates
attention on outputs whose coarse SQL and result signals indicate
risk, distinguishing the activation decision from the correction
itself.

\subsection{Candidate Coverage and Decision Quality}
\label{sec:candidate-coverage}

\paragraph{From candidate coverage to final accuracy.}
Table~\ref{tab:coverage-flow} separates absent solutions from lost
solutions at the record level. 60 of 759 covered records end with an
incorrect Full answer in the decision study, and 69 of 777 covered
records are lost in its retrieval-G2 arm. Neither Full cohort recovers
a record whose five candidates are all wrong; archived runs do show
such pool-empty rescues (six and two cases,
Table~\ref{tab:archived-judge}). Coverage fractions in
this record-level view describe the paired studies' own pools and
scoring protocol; they are not the Oracle percentages of
Table~\ref{tab:author-ablation-summary}, which count questions rather
than question--repetition records.

\begin{resulttable}
  \centering
  \setlength{\tabcolsep}{4pt}
  \begin{tabular}{lrr}
    \toprule
    Outcome & Decision & Retrieval G2 \\
    \midrule
    Correct candidate available & 759 & 777 \\
    Covered, final correct & 699 & 708 \\
    Covered, final incorrect & 60 & 69 \\
    Uncovered, final incorrect & 237 & 219 \\
    Uncovered, final correct & 0 & 0 \\
    \bottomrule
  \end{tabular}
  \caption{Candidate-to-Full outcomes in the paired studies, totaled
  over scored question--repetition records per study. Counts are
  per-study pool statistics and are not comparable to the
  question-level totals of
  Table~\ref{tab:author-ablation-summary}.}
  \label{tab:coverage-flow}
\end{resulttable}

Most final errors arise on uncovered questions: 237 of 297 decision
errors (79.8\%) and 219 of 288 retrieval-G2 errors (76.0\%) have no correct
candidate. These outcomes identify two routes for improvement:
knowledge-guided construction can create a correct alternative, and
adjudication can preserve the alternatives already available. Two-stage
retrieval strengthens the first route; the coverage-to-answer analysis
locates opportunities for the second.

\paragraph{Selection and semantic repair.}
Across the main ablation, the paired studies and the archived runs, the
Judge contributes to final answers in three distinct ways, each visible
in retained evidence. It can \emph{select a useful minority} when
execution groups disagree (Figure~\ref{fig:monthly-links-case}); it can
\emph{probe a disputed assumption} before deciding
(Section~\ref{sec:stage-reuse}); and it can \emph{reconstruct an
answer outside the pool} when every candidate shares a semantic error.
The first two are selection-level contributions; the third is what
Table~\ref{tab:archived-judge} records. It independently rescores two
five-candidate configurations with a \texttt{qwen3.7-max} Judge and no
active Reviewer. Within each run, the Judge, first-candidate and
majority views use identical saved candidates and a common evaluable
subset.

\begin{resulttable}
  \centering
  \small
  \setlength{\tabcolsep}{3pt}
  \begin{tabular}{lrrrrr}
    \toprule
    Candidate arm & $n$ & First & Majority & Judge & Rescues \\
    \midrule
    \texttt{qw3.6plus} & 497 & 385 & 386 & 397 & 6 \\
    \texttt{qw3.7max} & 496 & 392 & 397 & 402 & 2 \\
    \bottomrule
  \end{tabular}
  \caption{Correct-answer counts in two archived fixed-pool runs.
  Rescues have no correct original candidate. Arm names follow the
  archive; these are separate historical configurations, not current
  main-result repetitions.}
  \label{tab:archived-judge}
\end{resulttable}

The Judge gains twelve and ten correct answers over fixed-first
selection, and eleven and five over majority. Its six and two
pool-empty rescues are pure reconstructions: no original candidate is
correct, yet the final answer is. In Q00486, all five candidates
compute a card percentage using the sum of
mana costs as the denominator. The Judge instead divides the number of
qualifying cards by the total number of cards, yielding a correct result.
The correction changes the quantity being measured, not merely the SQL's
form. Selection, probing and reconstruction complement one another:
the first preserves the alternatives Candidates already found, the
last supplies an answer none of them contained.

Together, the selection gains and archived rescues demonstrate
complementary contributions from the Judge. It can choose a useful
alternative from existing evidence and reconstruct an answer when the
pool shares a semantic error. The complete release audit in
Appendix~\ref{app:archived-judge} reports the range of outcomes across
historical configurations, on their separate cohorts. The permission
ablation evaluates the policy that connects these capabilities to
individual decisions.

\paragraph{More candidates help coverage more than voting.}
We evaluate every nonempty subset of each five-slot pool, giving 31
subsets per question. For
each size $k$, scores average all $\binom{5}{k}$ subsets and both
repetitions. This keeps the candidate source fixed while varying the
number of candidates. The retrieval-study Oracle rises from 70.60\% at
$k=1$ to 78.01\% at
$k=5$, whereas majority EX changes from 70.60\% to 70.98\%
(Figure~\ref{fig:candidate-count}). The 7.41-point coverage gain yields
only a 0.38-point voting gain. Additional candidates frequently create
alternatives without making the largest agreement group more accurate.

\begin{resultfigure}
  \includegraphics[width=0.50\columnwidth]{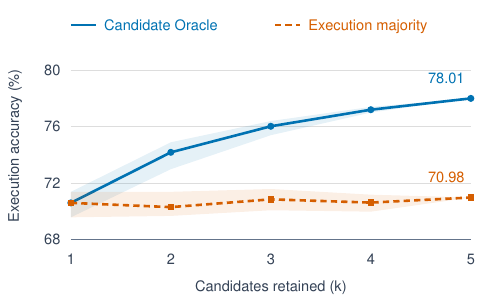}
  \caption{Fixed-pool candidate-count analysis in the paired retrieval
  study (its own pools and scoring; not the main-ablation scores of
  Table~\ref{tab:author-ablation-summary}). Lines
  average every subset of size $k$ and both repetitions. Shading shows
  the range across subset choices, not a confidence interval. Coverage
  increases much faster than execution-majority accuracy.}
  \label{fig:candidate-count}
\end{resultfigure}

\paragraph{Model mixtures expose the same selection gap.}
The archived 500-question candidate analysis extends the comparison to
DeepSeek and Qwen pools (Figure~\ref{fig:candidate-diversity}). Mix-6
combines the first three candidates from each model; Mix-10 uses all
five from each. Mix-6 covers more questions than either single-model
pool. Expanding to Mix-10 raises Oracle ExD from 78.00\% to 78.40\%,
but lowers majority ExD from 71.60\% to 70.80\%. These are recombinations
of the same source candidates, scored with the archive's ExD metric:
unordered, deduplicated execution results rounded to three decimal places.
Appendix~\ref{app:historical-results} records the source and tie policy.

\begin{resultfigure}
  \includegraphics[width=0.50\columnwidth]{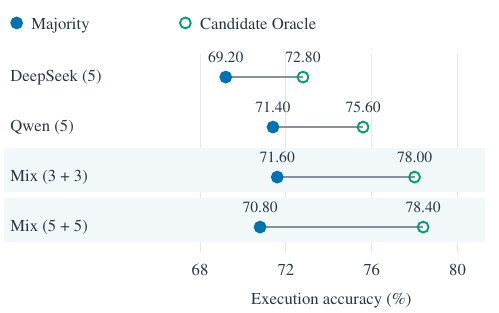}
  \caption{Candidate coverage and majority selection on an archived
  500-question cohort, scored with the archive's own ExD metric; these
  are not the main-ablation scores. Filled points show majority ExD; hollow points
  show Oracle. Mix-6 and Mix-10 reuse three or five candidates from each
  source model. The larger mixed pool has higher coverage and lower
  voting accuracy.}
  \label{fig:candidate-diversity}
\end{resultfigure}

Oracle rewards the presence of any correct candidate; majority selection
depends on the relative sizes of execution-result groups. An enlarged
pool can therefore contain more correct alternatives while making an
incorrect group more influential. The fixed-source subset analysis
isolates candidate count, whereas the mixed pools also change model
composition. Both analyses expose a decision problem beyond generating
additional alternatives: the final selector must preserve useful minority
answers when agreement favors a different interpretation.

\FloatBarrier

%% file: body/intervention-analysis.tex
\subsection{Intervention and Conditional Routing}
\label{sec:intervention-routing}

Specialized corrections address recurring ambiguities in aggregation,
joins and temporal conditions. The Reviewer comparison in
Section~\ref{sec:component-ablations} motivates a closer look at their
activation: which answers need intervention, and which semantic
conditions justify a change? A recorded edit and the archived mechanism
studies connect these questions to concrete SQL behavior.

\paragraph{Preserving the conditions that define an answer.}
In r1/Q00637, the Judge's correct query retrieves distinct tags for one
user's uncommented posts and filters \texttt{Tags IS NOT NULL}. The
Reviewer removes that filter after a \texttt{guard-suspect} trigger;
the saved score changes from correct to incorrect. Execution remains
valid, but the answer changes. The non-null filter defines the required
output in this question. This case identifies a useful condition for
review: establish a predicate's semantic role before replacing it.

\paragraph{Mechanism-specific repairs and regressions.}
The archived M1--M9 study applies dedicated semantic interventions on
216 matched questions, where the E1 baseline answers 159 correctly
(Figure~\ref{fig:historical-repairs}). None improves E1's total;
net changes range from zero to $-10$. Grain audit M1 repairs four answers
but breaks eight. Join contract M2 balances three repairs with three
regressions, while temporal solver M5 repairs two and breaks twelve.
Each of these interventions corrects some existing errors. Their net
outcomes depend on the correct answers they also change.

\begin{resultfigure}
  \includegraphics[width=0.55\columnwidth]{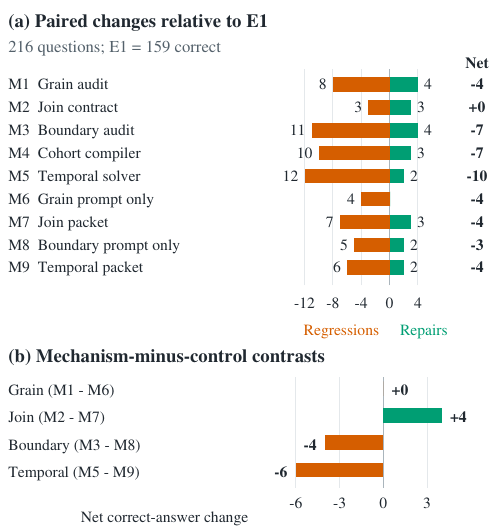}
  \caption{Mechanism interventions on 216 matched questions from an
  archived development cohort (E1: 159
  correct). (a) Repairs, regressions, and net change. (b) Contrasts against
  the corresponding MCP-free controls, including solver or gate additions.
  Paired counts distinguish a mechanism's corrections from the
  existing answers it changes.}
  \label{fig:historical-repairs}
\end{resultfigure}

The paired mechanism controls further separate useful tool-assisted
corrections from total accuracy. M1 and its prompt-only counterpart M6
both lose four answers relative to E1, although M1 adds four repairs
and four regressions. The join contrast M2--M7 gains four answers; the
boundary and temporal contrasts lose four and six. Additional tools and solver logic
thus have different effects across mechanisms, even within one matched
cohort. The archive and its execution exceptions are documented in
Appendix~\ref{app:historical-results}.

\paragraph{Conditional activation preserves the default path.}
\label{sec:historical-routing}
M10 uses a selective policy: it routes 21 questions to a grain branch
and 22 to a temporal branch, retaining 455 E1 default answers. Its
backtest combines archived outputs with cache-miss fills; R1 and R2
rerun the 43 routed questions with empty branch caches. The resulting
composites reach 377, 376 and 378 correct answers against
E1's 373 (Figure~\ref{fig:historical-routing}). The grain branch adds
four in every composite; the temporal branch adds zero, $-1$ and $+1$.
The positive aggregate change comes from the grain branch, while the
temporal branch varies around no net gain.

\begin{resultfigure}
  \includegraphics[width=0.50\columnwidth]{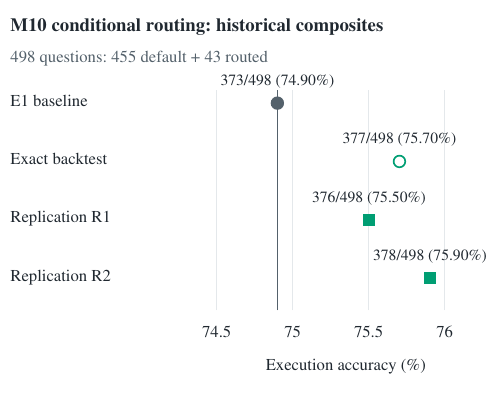}
  \caption{M10 routing composites from an archived development cohort.
  The hollow circle
  is the exact backtest; squares repeat the 43 routed questions.
  All retain 455 E1 default answers. The repeated grain-branch gain
  coexists with variation in the temporal branch. Repetitions measure
  output stability on the fixed routed subset.}
  \label{fig:historical-routing}
\end{resultfigure}

M10 makes the activation decision explicit: most questions retain their
existing answers, and each intervention is evaluated on the questions it
changes. Its repeated branch results describe this fixed routed subset,
rather than routing performance on unseen questions. Together with the
matched mechanism audit, they motivate evaluating branch selection
separately from branch execution. A useful correction procedure and a
useful rule for invoking it are distinct parts of an agent policy.
This raises a practical development question: which proposed correction
should become part of the next version? The following study applies
paired validation to that choice.

\FloatBarrier

%% file: body/evolution-analysis.tex
\subsection{Evidence-Guided System Evolution}
\label{sec:evolution}

\begin{figure}[!t]
  \centering
  \includegraphics[width=\textwidth]{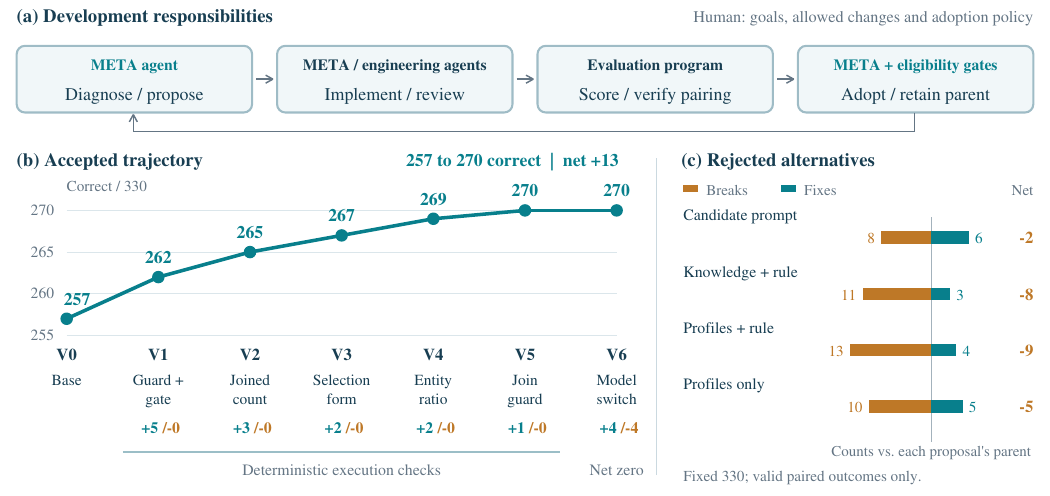}
  \caption{Evidence-guided development on the same 330 questions used
  for proposal development and version selection.
  (a) Agent and program responsibilities under human-set policies.
  (b) Accepted versions, annotated with the change and paired fixes/breaks.
  V1--V5 enable execution checks; V6 changes the Judge/Reviewer model.
  (c) Rejected proposals have both repairs and regressions; net losses
  compare each with its own parent. These development outcomes are
  separate from the 500-question main evaluation.}
  \label{fig:validated-evolution}
\end{figure}

The preceding analyses identify recurring semantic corrections.
We examine a development loop that turns these observations into
reusable reasoning policies
(Figure~\ref{fig:validated-evolution}). A development LLM agent, META,
examines saved failures and proposes changes. It can implement changes
directly or delegate coding and independent review to engineering agents.
Evaluation programs
execute the experiments and check scores, stage identities and eligibility.
META selects the next baseline within these checks, recording an explicit
rationale for below-threshold, non-negative changes. Human operators set
the goal, permitted changes and adoption policy, including model choices
and exceptions; eligible updates need no separate human approval.
These development roles are distinct from the online SQL agents.

The fixed development cohort contains 330 stratified questions from the
2024-06-27 Mini-Dev release and uses three \texttt{qwen3.7-plus}
Candidates. Children reuse compatible parent stages and are compared on
the same questions used to develop and select the changes.
Appendix~\ref{app:evolution} records the cohort, model configurations and
version history.

\paragraph{Narrow execution checks accumulate useful corrections.}
On the fixed 330 questions used for development and version selection,
the accepted sequence raises correct answers from 257 to 270,
or 77.88\% to 81.82\%. The first-to-last comparison contains 17 repairs
and four regressions, a net gain of thirteen answers
(3.94 percentage points). All of that net gain occurs in the first five
accepted changes, which enable deterministic checks after adjudication
and review.

These changes connect directly to the earlier failure analysis. The
joined-count check revisits an unrequested \texttt{COUNT(DISTINCT)};
the entity-percentage check tests a row-count denominator in place of
a measure-weighted ratio. An added-table guard can retain a supported
candidate answer when a later stage introduces an unsupported join.
Each has an explicit activation condition and checks the replacement
through execution. The first five parent--child comparisons have no
regressions. Their measured benefit comes from constraining particular
semantic edits.

\paragraph{Policy refinement and version adoption.}
Revised Candidate instructions repair six answers but break eight.
A knowledge-generator and subtotal-rule change repairs three and breaks
eleven; two column-profile variants also reduce the score. These proposals
are not adopted. The accepted execution checks instead constrain a
specific transformation under an explicit activation condition.

The final accepted model change illustrates a different outcome. It
repairs four answers and breaks four, leaving the total at 270.
The record adopts it under a separately identified non-negative-change
rule. The net development gain therefore comes from the execution
checks. They extend the earlier intervention analysis into an operating
procedure: diagnose a recurring semantic error, encode a targeted check,
and compare its effect with the parent policy before adoption.

\FloatBarrier

%% file: body/discussion.tex
\section{Discussion: Implications for Agent Design}
\label{sec:discussion}

RubikSQL-HB makes database knowledge available throughout reasoning, and
the evaluation connects knowledge acquisition to answer selection.
The example interface improves candidate coverage, and the Judge
improves selection from the available evidence. Continued access to
the workspace and database extends its role to investigating disputed
interpretations. The resulting design question is how to direct these
capabilities toward the semantic decisions that determine an answer.
The development study extends this analysis across versions by turning
recurring corrections into execution checks. Across both timescales,
knowledge is most useful when it supports a specific decision: choosing
an observation, selecting an answer or activating a correction.

\subsection{Knowledge Access Is Part of the Reasoning Policy}

The two-stage example interface exposes a choice that one-stage retrieval
performs implicitly: which reference SQL enters context. Its menu lets the
agent inspect the reference question before expanding a query. The measured
effect is a 1.0-point increase in final EX over one-stage
retrieval on the main ablation, alongside the coverage gains observed
in the record-level analyses.

Agents expand reference SQL in 29.7\% of question--repetition units. Discovery and expansion make the reference itself an object of
agent choice. Organizing this access changes the available solutions
even with the example index and ranking policy held fixed. The interface
therefore participates in the reasoning policy: it determines which
knowledge-selection decisions the agent can make.

Typed knowledge supports choices beyond example retrieval. In Q00532,
the Candidate inspects a field's documented meaning and observed timestamp
format before executing a year filter. The workspace exposes that
field-level evidence independently of complete reference queries.
Separating documentation from generated profiles lets the agent connect
a field's stated meaning to its observed contents. A database-specific
workspace makes both forms of evidence reusable across tasks while
preserving their source and scope.

\subsection{Adjudication with Continuing Evidence Access}

The Judge connects independent candidate search to the final answer.
The adjudicator can compare what queries mean alongside their execution
results, rather than relying on agreement alone.
The separate archived pool-empty repairs add a complementary role. In Q00486,
the Judge corrects a denominator shared by all five candidates. Selection
and reconstruction solve different problems, and each has a concrete
success mode in the retained evidence.

Continued tool access gives the Judge a way to investigate a concrete
disagreement. The denominator probe in Q00665 reveals the observed months;
the question determines the intended averaging period. A tool response
becomes useful through its relation to the disputed assumption. Here,
continued execution access lets the Judge check the interpretation behind
the competing answers. A useful next call distinguishes which entity is
counted, which population forms a denominator, or which field expresses
the requested concept. This provides a design criterion for investigation
beyond the availability of a tool.

The coverage analyses identify further opportunities for this decision
layer. In the fixed-pool subset analysis, additional candidates improve
coverage much more than voting accuracy. Mix-10 likewise has higher
Oracle coverage than Mix-6 but lower majority accuracy
(Figure~\ref{fig:candidate-diversity}). Full loses 60 covered records in
the decision study and 69 in retrieval G2. Retaining these available
solutions complements improvements to generation, since most remaining
errors arise without a correct candidate.

The current handoff carries SQL and bounded execution summaries, leaving
the Judge to reconstruct the evidence behind a Candidate's interpretation.
A useful extension would attach decisive observations to the SQL choices
they support, such as the entity counted or the field used for a business
concept. The Judge could then target unresolved differences while retaining
access to the original workspace. This would preserve independent search
without discarding evidence already acquired by the Candidates.

\subsection{Targeting Semantic Corrections}

The most actionable correction targets a semantic assumption: which
entity is counted, which population defines a denominator or which
condition determines an eligible row. The targeted retest repeatedly
exposes this structure. Patient-level counts benefit from
\texttt{DISTINCT}, whereas examination-record counts require preserving
multiplicity. In Q00637, the non-null filter is part of the intended
answer. These cases suggest that a correction policy should identify
the requested quantity before changing a familiar SQL pattern.

Inspection and revision can then have separate activation criteria.
The Reviewer gate concentrates errors in 35.7\% of scored records;
choosing a beneficial replacement remains a further decision, with
two repairs and two regressions under the measured gate. The Judge's
separate observation and revision permissions support this distinction:
an agent can investigate a disputed assumption and still retain the
existing query. The value of investigation lies in resolving the
assumption, whether or not the SQL changes.

Conditional routing applies the same principle to specialized procedures.
M10 retains 455 default answers and adds four correct answers through
the grain branch in each of its fixed-subset composites
(Figure~\ref{fig:historical-routing}). Its temporal branch varies around
zero. The branch-level analysis locates the useful correction and the
conditions under which it is invoked. Together with the mechanism
controls, it motivates a policy organized around specific semantic
decisions, with a supported default answer available throughout.

\subsection{From Execution Experience to Reusable Policies}

The development trajectory shows how semantic diagnosis can improve
subsequent runs. Its accepted checks convert recurring discrepancies
into explicit conditions on a later stage's output. On the fixed
330-question development and version-selection cohort, these checks
raise the correct count from 257 to 270. The model no longer has to
rediscover each correction in every conversation. META proposes the
mechanism, engineering agents can implement it, and the evaluation
program measures the change before adoption.

This suggests distinguishing database knowledge from execution policy.
A documented field meaning belongs in the workspace, available for
interpretation in context. A recurring transformation with a precise
activation condition can instead be implemented and tested as a runtime
check. The two representations serve different purposes: workspace
knowledge supports interpretation in context, while an execution check
encodes a specific transformation and its activation conditions.
Choosing the appropriate representation lets the system reuse experience
at the right level. RubikSQL-HB combines context-dependent reasoning
within a task with targeted policy refinement across versions.

Viewed together, these stages form a single knowledge lifecycle: the
workspace accumulates database knowledge before inference; Candidates
consult it and the live database while constructing SQL; the Judge
returns to the same evidence when adjudicating; and recurring
corrections are consolidated into execution checks that shape later
runs. Each stage of the evaluation examined one turn of this cycle.
Knowledge in the loop is the design principle that connects them:
keeping both stable background knowledge and live execution facts
available at every point where an agent must decide.

%% file: body/limitations.tex
\section{Limitations}
\label{sec:limitations}

The evaluation uses a public BIRD development subset that has influenced
system development. Training-example transfer uses the target questions
and hints while withholding target gold SQL. One transferred question
overlaps Q00915 in wording; its audited source is a training example with
different SQL. The comparisons characterize this development workflow
rather than performance on an unseen benchmark.
With 500 questions, one question equals 0.2 points, so main-ablation
gaps of one point or less are within question-level variation. The
paired intervals attached to the record-level analyses characterize
question-level variation.
The evolution cohort is also reused to select successive versions, so
its development gain does not estimate held-out improvement. That case
study evaluates a particular proposal sequence, without a controlled
comparison between META and alternative development workflows.

The two service-backed repetitions use recorded requested seeds;
provider-level determinism and immutable model versions remain unknown.
The adopted outputs include post hoc engineering recovery and reuse of
compatible stages, detailed in Appendix~\ref{app:run-record}. Paired
intervals describe question-level variation within these delivered runs.
The discordant-unit retest is selected from prior disagreements and
contains one fresh inference per arm and unit. It diagnoses variability
and semantic errors, not a new full-cohort treatment effect. Archived
Judge configurations use older questions and reference SQL; their
within-pool gains are analyzed separately from the current scores.

Stored profiles and rules can contain errors, whose individual accuracy
effects remain unmeasured. The failure analysis uses traceable SQL edits
and stage transitions; source-package error labels are agent annotations.
Question identifiers quoted in the case analyses (Q00665, Q00486,
Q00915, \emph{etc.}) are internal identifiers from the released
evaluation record.
EX measures agreement on a single database instance, where
different queries can agree accidentally. Other datasets and SQL dialects,
inference efficiency, and production reliability and security are outside
this answer-quality evaluation.

%% file: body/conclusion.tex
\section{Conclusion}

RubikSQL-HB turns distributed database knowledge into a resource that
agents can query while reasoning. HeavenBase supplies persistent workspace
infrastructure; RubikSQL-HB organizes database-specific knowledge and
connects it to an iterative Candidate process. Independent Candidates
construct and execute SQL, and downstream agents can reuse the same
interfaces to investigate completed answers.

The evaluation examines the knowledge interface and evidence-based
adjudication. On the 500-question BIRD Mini-Dev evaluation, the full
pipeline reaches 75.2\% execution accuracy against 70.4\% for
execution-majority voting, and the raw-knowledge condition, which also
regenerates candidates, is 3.0 points lower. A recorded execution probe shows the Judge resolving
a candidate disagreement, and separate archived runs demonstrate
pool-external repairs. Aggregation grain and denominator choice emerge
as concrete semantic questions that tool-assisted investigation and
revision can address. The ablations distinguish candidate creation,
answer selection and correction, giving each capability a specific
role in the system.

A separate development record extends this analysis to policy changes
between versions. On the fixed 330 questions used for development and
version selection, the adopted checks increase the correct count from
257 to 270. This gain comes from narrowly scoped execution checks in the
recorded proposal sequence. The checks encode recurring semantic
corrections with explicit activation conditions, carrying execution
experience into later runs.

The resulting design principle is to keep knowledge available at the
point of decision. A workspace supplies reusable evidence; Candidates
use it to construct alternatives; the Judge connects those alternatives
to a final answer through comparison and further investigation.
Across versions, targeted execution checks preserve useful corrections.
RubikSQL-HB connects these processes through a shared knowledge interface,
showing how persistent knowledge can support autonomous reasoning within
a task and the refinement of the policies used by subsequent tasks.

%% file: appendix/reproducibility.tex
\section{Reproducibility and Additional Results}
\label{app:reproducibility}

\subsection{Evaluation Record}
\label{app:run-record}

The decision and retrieval studies use the decision completion
package dated 2026-10-21 and the retrieval package's F20 adopted
outputs. Earlier F19d tables retained inside the retrieval narrative
are not the final results.
Both studies use the same ordered 500 questions, question text, hints,
database identifiers, and logical knowledge snapshot. Their independent
candidate generations use different seed schedules. Table~\ref{tab:runtime}
summarizes the configuration needed to interpret the comparisons.

\begin{resulttable}
  \centering
  \small
  \setlength{\tabcolsep}{3pt}
  \begin{tabularx}{\columnwidth}{lX}
    \toprule
    Item & Setting \\
    \midrule
    Reference code & recorded commit, with recorded adapters \\
    Candidate model & \texttt{qwen3.7-plus}, five fixed slots \\
    Judge (Max) / (Plus) & \texttt{qwen3.8-max} / \texttt{qwen3.7-plus} \\
    Reviewer & Reviewer v2, following the Judge model \\
    Embedding & \texttt{text-embedding-v3}, 1024 dimensions \\
    Paired-study R1/R2 & Candidates 1--5 / 101--105; Judge 42/142;
      Reviewer 45/145 (decision); 201--205 / 301--305; Judge 242/342;
      Reviewer 245/345 (retrieval) \\
    EX comparator & Exact, unordered sets of positional rows;
      no numerical rounding \\
    Gold precheck & 30 CPU seconds per query \\
    Uncertainty & 10,000 qid-clustered bootstrap draws,
      seed 20261015 \\
    \bottomrule
  \end{tabularx}
  \caption{Evaluation configuration. Requested seeds do not imply
  provider-guaranteed deterministic generation.}
  \label{tab:runtime}
\end{resulttable}

The delivered views contain 8,000 decision results and 12,000 retrieval
results, with 5,000 and 15,000 adopted candidate slots, respectively.
Each condition and repetition covers all 500 question IDs without
duplicates. The current analysis independently recomputes their summary
scores, mean paired differences, and decision and retrieval bootstrap
intervals from the supplied records. It also checks the 31-subset Oracle
records. The recomputation uses the retained per-question execution scores.

The accompanying compact evidence exports retain summaries, paired
statistics, database breakdowns, source archive digests, and the figure
regeneration script in \path{experiment-evidence/20261022/}.
The complete delivered archives retain the more extensive stage records.
The retrieval ZIP is a paper-oriented export: some source-artifact and
SQL-cache references resolve in its producer's complete archive rather
than inside that ZIP.

\paragraph{Knowledge preparation.}
The approved snapshot uses training-source examples transferred to the
target schemas. Per-question top-four source selection uses the fixed 500
target questions and hints; the transfer model receives source pairs and
the target schema, not target gold answers. The audit records one
approved question-text overlap at Q00915, with a training source and a
different SQL query. Historical E1 used a different construction scope
based on all 1,534 development questions, so its knowledge state is not an
exact replica of the current one.

The frozen profiling code can join unrelated tables on identically named
primary keys and interpret key equality as evidence of a shared entity,
although the join condition already imposes that equality. The diagnosis
also records overbroad rules. Documented meanings and generated profiles
remain distinct sources; the individual accuracy effects of these stored
inferences were not isolated in the delivered comparisons.

\paragraph{Recovery and final-result selection.}
The packages preserve earlier attempts and adopt corrected outputs under
their documented engineering-recovery policies. The decision completion
reuses all candidate slots and input-compatible stages while repairing
Judge execution. It includes a separately identified Q00062 recovery.
The original freeze discrepancies remain recorded rather than being
retroactively rewritten. The retrieval F20 update adopts 411 affected
question--condition--repetition units after candidate transport and
execution recovery, followed by their Judge and applicable Reviewer
stages. Its final non-hash validation records no failed checks, but does
not revalidate the complete earlier history. Recovery policies were
amended after initial execution; the reported comparisons describe the
final adopted runs.

The decision package retains three newly recorded unknown-delivery
requests in J01: r1/Q01392, r2/Q01011, and r2/Q00062. Its request-closure
check therefore reports \texttt{PARTIAL}, although all scored result
views are present. These unknown requests are not additional missing
score rows. J01 has 147 and 122 fallback outputs in R1 and R2. We retain
those outputs in its EX and interpret the contrast as the behavior of
the implemented policy, rather than omitting difficult responses.
Thirteen decision and 27 retrieval candidate slots have terminal failed
statuses. Full retains two final upstream failures in the decision study
and one final failure in retrieval G2; evaluable failed predictions score
zero.

\paragraph{Illustrative knowledge trace.}
\label{app:knowledge-case}
The Candidate example in Section~\ref{sec:candidate-loop} uses retrieval
F20 r2/G2/Q00532, slot zero, from the delivered
\path{sample-arm-records.zip}. The retained tool transcript includes
\texttt{get\_db\_info}, the nonempty \texttt{get\_col\_info} response for
\texttt{users.CreationDate}, and successful execution and base-database
verification of the displayed SQL. The adopted retrieval export marks
that Candidate and its Full result correct. The case description uses
observable calls, returned field metadata and saved scores, without
re-execution or a claim that the lookup was necessary for correctness.
Exact archive paths and digests are recorded in
\path{experiment-evidence/20261022/knowledge-case-provenance.md}.

\paragraph{Illustrative execution trace.}
\label{app:execution-case}
Figure~\ref{fig:monthly-links-case} uses Q00665 from the
\texttt{ablation-20261008} development run, attempt 1. The retained
trajectory contains the five candidate results and the monthly-count
probe; read-only re-execution on the manifest-matching database reproduces
the displayed values and the Judge's reference match. The run removed the
hint, its column-information lookup failed, and example retrieval returned
no examples. The case records execution-based investigation of a
denominator disagreement. The schema shown in the figure is reconstructed
for explanation. Trace and database digests are retained in
\path{figures/method-figures-provenance.md}.

\subsection{Paired Effects and Repetition Results}

The paired studies in Sections~\ref{sec:component-ablations}
and~\ref{sec:candidate-coverage} are two-repetition studies on their
own fixed candidate pools and an earlier scoring protocol. They are
reported as mechanism analyses with paired bootstrap intervals; their
absolute scores and contrasts are separate from the main ablation of
Table~\ref{tab:author-ablation-summary} and must not be
compared with it row by row.

\begin{resulttable}
  \centering
  \small
  \setlength{\tabcolsep}{3pt}
  \input{experiment-evidence/20261022/paired-effects.tex}
  \caption{Paired effects in the two-repetition paired decision study
  (its own pools and scoring protocol; separate from the main
  ablation). Differences are percentage points, with qid-clustered
  95\% intervals. Both repetitions move together when a question is
  resampled.}
  \label{tab:paired-effects}
\end{resulttable}

The retrieval study's R1 Full scores are 72.09\%, 70.48\%, and 71.69\%
for G0, G1, and G2; R2 scores are 70.88\%, 72.29\%, and 70.48\%.
The G2--G1 final contrast therefore changes sign across repetitions.
The omitted fixed-first means are 70.58\%, 70.38\%, and 69.58\%.

For Full, G1--G0 is $-0.10$ points with interval $[-1.61,1.41]$;
G2--G0 is $-0.40$ with interval $[-1.71,0.90]$; and G2--G1 is
$-0.30$ with interval $[-1.71,1.10]$. The Oracle contrast G2--G1 is
$+2.01$ with interval $[0.70,3.51]$. These comparisons use the same
gold-valid question set and do not use historical absolute scores as
a stopping or selection criterion.

The selection-only J00--majority contrast in the decision study is
recomputed from the delivered per-question rows using the same
qid-clustered bootstrap: 23 wins, seven losses, $+1.61$ percentage
points, and interval $[0.40,2.91]$. J00--first is $+1.41$ points with
interval $[-0.20,3.01]$. These are comparisons of retained outputs on
the original cohort, not additional model runs.

\subsection{Discordant-Unit Judge Retest}
\label{app:judge-retest}

The subsequently delivered Experiment~1 new-results ZIP contains
\path{analysis-v2.md}, \path{difference-explanation.md}, and
\path{per-question-new.md}. Its 46 units are question--repetition pairs
on which at least two original Judge arms disagree; they cover 37
distinct questions. Four arms are rerun on each unit, producing 184
reported outputs. The report records matching original candidate-pool
and base-input hashes for all 184. Compact exports and source digests
are retained in \path{experiment-evidence/20261024/}.

The delivered per-unit score table reproduces the counts in
Table~\ref{tab:judge-retest} and the J11--J00 transition counts. The
producer additionally reports closed requests and unchanged knowledge
and database hashes. The underlying HTTP, worker and verification JSON
files named in those documents are not included in this three-file ZIP;
their protocol claims are therefore distinguished from independently
recomputed score-table counts.

J01's retest has 46 fallback labels: 43 final SQL strings match a
candidate, while three pool-external queries are incorrect. J11 has
five pool-external queries, four correct. Every selected pool already
contains a correct candidate, so these four are not pool-empty rescues.
The producer records SQL execution calls on 30/46 J10 units and 34/46
J11 units. J00's 37 requests for unavailable observation tools are
reported as denied without database information; tools-off still
includes the candidate execution previews supplied initially.

The original J11 outputs include substantially more historical reuse
than the other arms: 42 of these 46 units, versus five per other arm.
The retest places all four arms in one new inference window. Requested
seeds do not guarantee deterministic service responses, and the
selected subset is susceptible to regression toward agreement. Its
new outputs are not spliced into the old cohort to construct an updated
500-question score.

\subsection{Database-Level Outcomes}

\begin{resulttable}
  \centering
  \small
  \setlength{\tabcolsep}{3pt}
  \input{experiment-evidence/20261022/per-database.tex}
  \caption{Mean Full EX by database. A is the decision cohort; B
  denotes retrieval G0/G1/G2. Each column averages its own two
  repetitions. A and B use separate generated candidate pools.}
  \label{tab:database-results}
\end{resulttable}

Database outcomes vary substantially. Different example strategies lead
on different databases, and similar aggregate scores conceal opposite
changes across databases.

\subsection{Archived Result Provenance}
\label{app:historical-results}

The archived intervention figures use commit
\texttt{e3be71900fc6} of the RubikSQL-HB
\href{https://github.com/edsml-jz2321/RubikSQL-HB/tree/e3be71900fc65510c8ccdd1bfc49873dbe814d1d/experiments/m10_followup}{M10 follow-up directory}.
\path{FOLLOWUP_AUDIT.md} supplies the M1--M9 matched 216-question counts;
\path{M10_FINAL_REPORT.md} supplies the final v2 composites and routed
replications. M1 and M3 each include a final timeout. M10 reuses 455 E1
default outputs and reruns only 43 routed questions; incomplete candidate
trajectories for Q01239 do not contribute to the recorded gain.
The three composites share these default outputs.

The candidate-mixture figure uses
\href{https://github.com/edsml-jz2321/RubikSQL-HB/blob/4198c5d8305fbe3327b8c9062e244198a1a132cf/eval_results/eval_ruleDevelopVersion/majority_vote_analysis.json}{\texttt{majority\_vote\_analysis.json}}
at commit \texttt{4198c5d8305f}. Its 500-question raw Mini-Dev analysis
compares unordered, deduplicated execution results rounded to three
decimal places. Ties favor fewer output columns, followed by a seeded
choice. Mix-6 and Mix-10 recombine the first three or all five source
candidates. These archived metrics are kept separate from the current
strict EX results.

\subsection{Archived Judge Comparisons}
\label{app:archived-judge}

The Judge audit uses the repository's
\href{https://github.com/edsml-jz2321/RubikSQL-HB/releases/tag/snapshot-20260725}{snapshot-20260725}
and
\href{https://github.com/edsml-jz2321/RubikSQL-HB/releases/tag/storage-archive-20261013}{storage-archive-20261013}
release assets. Downloaded bytes match their published SHA-256 digests.
The two positive candidate configurations in
Table~\ref{tab:archived-judge} come from
\path{eval_result_candidate/qw3.7max/} with subdirectories
\path{qw3.6plus} and \path{qw3.7max}. Table~\ref{tab:release-audit}
reports the broader set of audited configurations.

\begin{resulttable}
  \centering
  \scriptsize
  \setlength{\tabcolsep}{2.5pt}
  \input{experiment-evidence/20261024/release-audit.tex}
  \caption{Independently rescored archived views: common evaluable
  denominator and correct counts for first, majority and Judge.
  Official primary uses the pre-Reviewer answer. Configurations are
  separate saved runs, not exchangeable repetitions.}
  \label{tab:release-audit}
\end{resulttable}

The audit executes saved SQL read-only and compares unordered,
deduplicated positional rows without rounding. A four-second execution
guard and a 250,000-row cap apply uniformly. A gold or prediction
timeout makes a record unavailable for the complete within-pool
comparison; invalid SQL remains a failed prediction. These rules
produce the explicitly reported denominator for each row. Majority
selects the largest successful execution group, resolving ties by
column count and then a deterministic question-ID-seeded choice.
Tie sensitivity preserves the positive sign in both highlighted runs.

For \texttt{qw3.6plus}, the excluded IDs are Q00518 and Q00701
(gold timeout) and Q00530 (prediction timeout). For \texttt{qw3.7max},
they are Q00518, Q00701, Q00408 and Q00637. The old 500-question
candidate runs differ from the cleaned reference in 158 gold SQL
records. Their absolute accuracy is consequently not a replication of
the main experiment. Comparisons are made within each saved pool.

The first configuration has six pool-empty rescues: Q00017, Q00486,
Q00726, Q00743, Q00894 and Q00989; the second has Q00634 and Q00989.
Q00486's final denominator is \texttt{COUNT(*)}, matching the reference
\texttt{COUNT(id)} on the archived database, whereas all candidates use
a sum of mana costs. Saved SQL and execution outcomes establish the
repair, but do not isolate a particular tool call as its cause.

The J1, J2 and J3 rows compare saved Judge variants. The putative older
73.40\% baseline shares only 60 of 499 normalized candidate pools with
J1, so subtracting it from J1's stored 74.95\% is not a validated
fixed-pool prompt effect. Official primary's post-Reviewer result is
also kept distinct: it has 361 correct records, versus 357 before
review. Source inventories, rescored counts and complete paired flips
are recorded in \path{experiment-evidence/20261024/}.

\subsection{Versioned Development Record}
\label{app:evolution}

The development analysis uses the authority branch
\texttt{authority/epoch-20261013-m1013}, pinned at
\texttt{9964b0ad1392}, and the accompanying code state
\texttt{47506ef9477a}. Proposal records identify the parent release,
intended change and falsification conditions. Outcome records retain
per-question parent and child scores, coverage and integrity checks;
separate decisions identify the release adopted as the next baseline.
The pinned operating guide assigns investigation and proposal authorship
to the META development agent, with engineering agents available for
code changes and independent review; META can also modify tools directly.
Programs execute evaluation and enforce integrity.
The recorded accepted decisions name the META development agent.
Ordinary promotion is outcome-gated; below-threshold adoption additionally
requires META's explicit evidence-backed rationale. The user authorizes
the scope and changes to that policy, including the model switch and
non-negative-change rule. Verdict overrides and rollback are human-only
operations. These roles and authority boundaries are retained with exact
source excerpts in \path{experiment-evidence/evolution-20261024/role-provenance.json}.
The analysis recomputes counts from question-level rows, verifies the
same 330 unique IDs in each comparison and checks that each accepted
child's scores equal the next accepted parent's scores.

\begin{resulttable}
  \small
  \setlength{\tabcolsep}{3pt}
  \begin{tabularx}{\columnwidth}{lXrrr}
    \toprule
    Version & Accepted change & Correct & Fix & Break \\
    \midrule
    V0 & Initial parent & 257 & --- & --- \\
    V1 & Consensus guard + submission gate & 262 & 5 & 0 \\
    V2 & Joined-count correction & 265 & 3 & 0 \\
    V3 & Selection normalization & 267 & 2 & 0 \\
    V4 & Entity-percentage count & 269 & 2 & 0 \\
    V5 & Strict added-table guard & 270 & 1 & 0 \\
    V6 & Judge / Reviewer model change & 270 & 4 & 4 \\
    \bottomrule
  \end{tabularx}
  \caption{Accepted development versions on 330 questions. Fix and
  break counts compare each version with its immediate parent.}
  \label{tab:evolution-versions}
\end{resulttable}

The question set derives from the 2024-06-27 raw Mini-Dev 500. Historical
timeout screening removes 82 questions, and gold execution removes
Q00518, leaving 417 eligible questions. Fixed database and difficulty
stratification selects 330. This older question/reference snapshot and
the three-Candidate configuration differ from the cleaned 500-question
main study. The source records include the deterministic selection rule
and question-set digest.
Candidates remain \texttt{qwen3.7-plus}; the Judge and Reviewer use
\texttt{qwen3.7-max} through V5 and \texttt{qwen3.8-max} in V6.

V1--V5 enable post-selection hooks. Their conditions cover departures
from candidate consensus and specific counting, arithmetic, date or
selection forms. These conditions were developed using the cohort's
errors and reference conventions. They are operational heuristics tested
on this cohort, not general SQL equivalence rules. Candidate-stage reuse
limits recomputation to affected stages; it does not turn the sequence
into six independent full-pipeline repetitions.

The ordinary promotion rule requires at least two net correct answers
and 0.5 percentage points. V5 and V6 use a separately recorded
non-negative-change adoption rule with an explicit rationale. V6's
decision also cites additional analyses, whose raw runs are not in
the authority export. The report uses only its verified counted outcome:
270 versus 270, with four repairs and four regressions. The accepted
first-to-last comparison has seventeen repairs and four regressions.

The valid rejected Candidate-instruction, knowledge-plus-rule,
profile-plus-rule and profile-only outcomes have respective
repair/regression counts of 6/8, 3/11, 4/13 and 5/10. These compare
with their own parent versions. A Judge/Reviewer prompt trial and an
earlier model-change attempt fail integrity verification and do not enter
the figure. Three later prompt trials also fail integrity checks during
an importer defect that disrupts tool calls. The subsequent code repair
does not supply a valid replacement cohort result in the pinned record.
Their failed predictions are therefore not treated as algorithmic
regressions or as evidence that the prompt changes are ineffective.

Compact records retain source Git blob IDs and SHA-256 digests in
\path{experiment-evidence/evolution-20261024/source-records.json}.
The adjacent summary contains recomputed transitions, accepted decisions
and excluded outcomes. Running
\path{experiment-evidence/integrate_evolution_20261024.py} regenerates
the summary and vector figure from these records without inference
or SQL execution. The figure's V0--V6 labels distinguish accepted
versions from the retrieval study's G0--G2 conditions.

%% file: experiment-evidence/20261022/paired-effects.tex
\begin{tabular}{lrr}
\toprule
Paired contrast & $\Delta$ EX & 95\% interval \\
\midrule
Full minus first & +0.50 & [-1.10, +2.11] \\
Full minus majority & +0.70 & [-0.50, +2.01] \\
Tools, selection only & -0.50 & [-1.51, +0.50] \\
Tools, revision allowed & +0.40 & [-0.80, +1.61] \\
Revision, tools off & -1.31 & [-2.51, -0.10] \\
Revision, tools on & -0.40 & [-1.41, +0.50] \\
Review all minus no review & +0.30 & [-0.20, +0.90] \\
Gated minus no review & +0.00 & [-0.40, +0.40] \\
Tools--revision interaction & +0.90 & [-0.70, +2.51] \\
\bottomrule
\end{tabular}

%% file: experiment-evidence/20261022/per-database.tex
\begin{tabular}{lrrrr}
\toprule
Database & A Full & B G0 & B G1 & B G2 \\
\midrule
California schools & 41.67 & 45.00 & 45.00 & 41.67 \\
Card games & 67.65 & 69.61 & 69.61 & 69.61 \\
Codebase community & 80.21 & 82.29 & 79.17 & 78.12 \\
Debit cards & 70.00 & 75.00 & 73.33 & 71.67 \\
European football & 73.53 & 73.53 & 75.49 & 72.55 \\
Financial & 42.19 & 43.75 & 46.88 & 48.44 \\
Formula 1 & 71.97 & 75.00 & 74.24 & 74.24 \\
Student club & 93.75 & 93.75 & 93.75 & 93.75 \\
Superhero & 93.27 & 94.23 & 93.27 & 95.19 \\
Thrombosis & 57.00 & 55.00 & 55.00 & 57.00 \\
Toxicology & 56.25 & 56.25 & 57.50 & 56.25 \\
\bottomrule
\end{tabular}

%% file: experiment-evidence/20261024/release-audit.tex
\begin{tabular}{lrrrrr}
\toprule
Archived view & $n$ & First & Majority & Judge & $\Delta$ count \\
\midrule
Candidate qw3.6plus & 497 & 385 & 386 & 397 & +11 \\
Candidate qw3.7max & 496 & 392 & 397 & 402 & +5 \\
E1 & 493 & 357 & 363 & 363 & +0 \\
J1 literal & 495 & 364 & 364 & 368 & +4 \\
J2 votes3 & 495 & 364 & 364 & 361 & -3 \\
J3 literal+votes3 & 495 & 364 & 364 & 365 & +1 \\
JM1a & 490 & 356 & 358 & 363 & +5 \\
HB500 off & 492 & 373 & 375 & 375 & +0 \\
Pipeline hidden500 & 495 & 353 & 363 & 352 & -11 \\
Final D old500 & 495 & 390 & 389 & 394 & +5 \\
T1 Qwen & 493 & 359 & 363 & 361 & -2 \\
Official secondary & 496 & 357 & 359 & 363 & +4 \\
Official primary (pre) & 496 & 349 & 353 & 357 & +4 \\
\bottomrule
\end{tabular}